\documentclass[runningheads]{llncs}
\usepackage{graphicx}
\usepackage{amsmath,amssymb} 
\usepackage{color}
\usepackage[width=122mm,left=12mm,paperwidth=146mm,height=193mm,top=12mm,paperheight=217mm]{geometry}

\usepackage{cite}
\usepackage{times}
\usepackage{epsfig}
\usepackage{xcolor}
\usepackage{multirow}
\usepackage[normalem]{ulem}
\usepackage{kotex}
\usepackage{booktabs}

\newcommand{\mb}[1]{\mathbf{#1}}

\newcommand{\rvauthcommands}[4]{%
\expandafter\newcommand\csname #1rev\endcsname{#4} 
\expandafter\newcommand\csname #1\endcsname[2][0]{\ifnum##1>\csname #1rev\endcsname\textcolor{#3}{##2}\else##2\fi} 
\expandafter\newcommand\csname #1t\endcsname[2][0]{\ifnum##1>\csname #1rev\endcsname\textcolor{#3}{\sout{##2}}\fi} 
\expandafter\newcommand\csname #1r\endcsname[3][0]{\ifnum##1>\csname #1rev\endcsname\textcolor{#3}{\sout{##2}##3}\else##3\fi} 
\expandafter\newcommand\csname #1c\endcsname[2][0]{\ifnum##1>\csname #1rev\endcsname\textcolor{#3}{/\uline{##2}/}\fi} 
}

\rvauthcommands{ml}{mlee}{red}{5} 
\rvauthcommands{gh}{geonho}{blue}{4} 
\rvauthcommands{sh}{songhwai}{green}{1} 

\begin{document}
\pagestyle{headings}
\mainmatter
\def\ECCV18SubNumber{2297}  

\title{Deep Pose Consensus Networks} 

\titlerunning{Deep Pose Consensus Networks}

\authorrunning{Geonho Cha, Minsik Lee, Jungchan Cho, and Songhwai Oh}

\author{Geonho Cha$^1$, Minsik Lee$^2$, Jungchan Cho$^3$, and Songhwai Oh$^1$}
\institute{$^1$Department of ECE, ASRI, Seoul National University, Korea\\
$^2$Division of EE, Hanyang University, Korea\\
$^3$Department of Software, Gachon University, Korea}

\maketitle

\begin{abstract}
In this paper, we address the problem of estimating \gh[3]{a} 3D human pose from a single image, which is important but difficult to solve due to many reasons, such as self-occlusions, wild appearance changes, and inherent ambiguities of 3D estimation from a 2D cue.
These difficulties make the problem ill-posed, which \ml{have become requiring} increasingly complex estimator\ml{s} to enhance the performance.
On the other hand, most existing methods \mlr{have handled the}{try to handle this} problem based on a single complex estimator, which might not be good solutions.
In this paper, \ml{to resolve this issue,} we \mlt{first }propose a \ghr[1]{consensus-based}{multiple-\ml[3]{partial-}hypothes\ml[3]{is}-based} framework for the problem of estimating 3D human pose from a single image\ml{, which can be \ghr[2]{trained}{fine-tuned} in an end-to-end fashion}.
We \ml{first} select \ml{several} joint groups from a human joint model using the proposed \gh[1]{sampling} scheme, and estimate the 3D poses of each joint group separately \ml{based on deep neural networks}.
After that, they are aggregated to \ml{obtain} the final 3D poses using the proposed robust optimization formula.
\mlr{The model could be trained in an end-to-end fashion}{The overall procedure \gh[1]{can be} \ghr[2]{trained}{fine-tuned} in an end-to-end fashion}, resulting in better performance.
In the experiments, the proposed framework shows the state-of-the-art performances on popular benchmark data sets, namely Human3.6M and HumanEva, which demonstrate the effectiveness of the proposed framework.
\keywords{3D human pose estimation, \ghr[1]{\ml{c}onsensus}{multiple-\ml[3]{partial-}hypothes\gh[2]{is}}-based model, \ml{a}rticulated pose estimation}
\end{abstract}

\begin{figure*}[t]
    \centering
    \includegraphics[height=3.2cm]{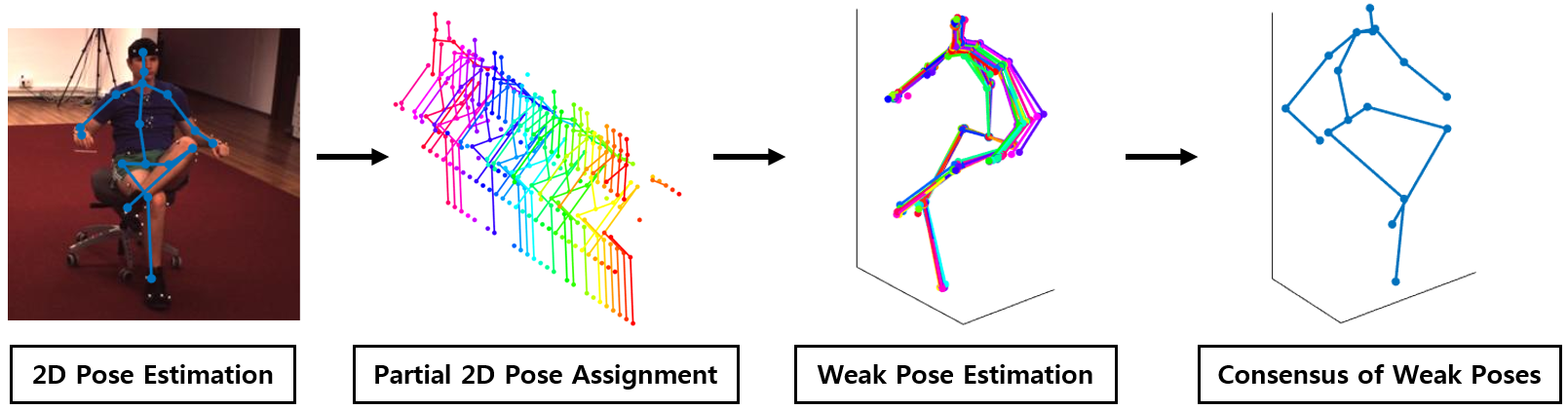}
    \caption{An overview of the proposed method. Given \ml{an} RGB image, the 2D pose is estimated. After that, each \gh[1]{predetermined} joint group selects \mlr{each}{a} partial 2D pose input and lifts the partial 2D pose to \ml{a} partial 3D pose (Different color\ml{s} mean\mlt{s} different joint group\ml{s}). Finally, \ml{the} weak reconstructions are aggregated to \ml{form} the final whole 3D pose.}
    \label{fig:overview}
\end{figure*}

\section{Introduction}
\mlr{We handle}{In this paper, we deal with} the problem of estimating \gh[3]{a} 3D human pose from a single image, whose objective is to infer the 3D coordinates of the whole body joints given an image.
The human pose, which is one of the most valuable information to understand visual data, can be used in many applications, such as human computer interaction, surveilance, augmented reality, video analysis, to name a few.
\mlr{According}{Due} to its importance, human pose estimation has been actively researched in computer vision for the last decade \sh[1]{\cite{bourdev:poselets:cvpr09, yang:articulated:pami13, toshev:deeppose:cvpr14, wei:cpm:cvpr16, newell:stakced:eccv16, chen:synthesizing:3dv16, rogez:mocap:nips16, park:3dhumanpose:eccv16, pavlakos:coarse:arxiv16, zhou:sparseness:cvpr16, yasin:dualsource:cvpr16, chang:consistency:17,  chen:3dpose2dposepmatching:cvpr17, martinez:asimply:iccv17}}.

In the early days, \mlr{they}{people} focused on 2D human pose estimation \cite{bourdev:poselets:cvpr09, yang:articulated:pami13, toshev:deeppose:cvpr14, wei:cpm:cvpr16, newell:stakced:eccv16} although it is less informative compared to the 3D case.
\gh[3]{This is} because \gh[1]{obtaining reasonable 2D estimation results was difficult} due to \ml[1]{many reasons such as} self-occlusions, wild appearance changes, and the high degrees-of-freedom of poses.
Early works \ml{on} 2D pose estimation \cite{bourdev:poselets:cvpr09, yang:articulated:pami13} utilized hand-crafted features like histogram of oriented gradients (HOG) \cite{dalal:hog:cvpr05}, resulting in \ght[3]{a }bad performance.
In the meantime, advances in \ml{c}onvolutional \ml{n}eural \ml{net}works (CNN) have made breakthroughs in several computer vision applications, including the 2D human pose estimation problem \cite{toshev:deeppose:cvpr14, wei:cpm:cvpr16, newell:stakced:eccv16}.
By applying CNN to \mlt{the }2D human pose estimation, \mlr{we could}{one can} learn \ml[3]{rich} features\mlt[1]{ in an end-to-end fashion}\mlt[3]{, which are discriminative\ght[1]{to appearance changes} and} \ml[3]{that are} robust to self-occlusions\ml[1]{, in an end-to-end fashion}.
As a result, \mlr{we could}{one can} obtain reasonable 2D estimation results.
Here, the performance\mlt{s} of the CNN-based estimators \ghr[3]{have}{has} \mlt{been }increased as the complexity of the CNN \mlt{are }increased.

Recent advances in \mlt{the }2D human pose estimation have promoted the research of 3D human pose estimation.
\ml{On the other hand, most of \ghr[1]{these}{the}} previous approaches have tackled the problem based on \mlr[4]{a single complex estimator\gh[3]{, i.e., the 3D pose is estimated using a hypothesis}}{an (possibly complicated) estimator that yields a single accurate hypothesis}.
\mlr{However}{Unfortunately}, the problem is highly ill-posed due to many reasons such as \ghr[3]{self-occlusions, wild appearance changes, and inherent ambiguities of the 3D estimation}{inherent ambiguities of the 3D estimation and self-occlusions}.
\ml{Hence, t}he complexity of \mlr{the single}{a 3D human pose} estimator has been \ml{continuously} increased to improve \mlt{the }accuracy, which would need \ml[1]{more and} more training samples\mlt{ that are difficult to collect}.
\ml{However, c}ompared to the case of 2D pose estimation, \ml{it is more difficult to collect 3D training samples because} a specialized motion capture system is needed\mlt{ to collect the training samples}.
In this light, a single\ml{-}estimator\ml{-}based approach\mlt{es} may not be \ml{the best} solution\mlt{s}.

\mlr{Instead of}{Unlike the} conventional schemes, we \mlr{could}{can} utilize \ml[1]{multiple hypotheses from} \ml{many different estimators}.
\ml{It is well-known that m}any complex problems have been successfully handled based on \ml{this} strategy in computer vision \cite{viola:robust:ijcv04, breiman:random:ml01}.
\mlr{In this way}{In this strategy}, the final estimation \ml{can be} obtained \ml{robustly} by aggregating many \ml{``weak''} estimations.
Here, some of the weak estimations could be bad, and the weak estimator is usually much simpler than the one based on a single estimator.

The complexity of \mlt[3]{each }weak estimator\ml[3]{s can} be further reduced \mlr[3]{by incorporating}{if we make them estimate} \ghr[1]{a part-based approach}{\mlt[3]{multiple }\emph{partial} hypotheses}, i.e., each weak estimator estimate\ml{s} the pose of some partial joints instead of the pose of the whole joints.
Actually, the structure of \ml{a} human body is appropriate to adopt \ml[3]{a} \ghr[1]{part-based}{multiple-partial-hypothes\ml[3]{is}-based} scheme\ml{,} because the human body is composed of four limbs that could move \ghr[3]{independently}{freely}\ml{. Here, the} four limbs are \ml{the} left leg, \ml{the} right leg, \ml{the} left arm, and \ml{the} right arm.
If \mlt{we always move }each limb \ml{is always moved} independently, the degrees of freedom \ml{for} modeling each limb separately would be lower than that of modeling the whole body joints simultaneously.
Even if \ml{this} assumption is not strictly true, we could expect that the degrees of freedom could be reduced by finding a meaningful partial joint groups.

\shr[1]{However}{On the other hand}, adopting the \ghr[1]{part-based}{multiple-partial-hypothes\ml[3]{is}-based} approach to the 3D human pose estimation is not a simple problem.
To achieve this, two issues need to be considered.
First, we need a proper method of selecting joint groups.
Even though there is a chance to reduce the complexity with a \ghr[1]{part-based}{multiple-partial-hypothes\ml[3]{is}-based} approach, improper joint groups \mlr{could}{can} ruin the final estimation result.
\gh[4]{To resolve this, we propose a joint group selection scheme which puts joints that are interdependent to the same joint group.}
The second issue is about aggregating the weak pose estimations.
This is also \ml{a} difficult problem because \mlr{of}{there are} translation ambiguities between partial poses.
Most of \mlt{the }regression models assume \mlr{zero mean of the data}{that the data has a zero mean}, and these removed translation components should be revealed before the aggregation. 
\gh[4]{We propose a robust optimization formula which aggregates the weak pose estimations \ml[5]{while} resolving the translation ambiguity issue.}

\gh[1]{\mlt[1]{Actually, }\ml[1]{The} main idea of this paper was inspired by \cite{lee:consensus:cvpr16}, which applied the concept of \ml[1]{multiple \gh[1]{partial} hypotheses} to the non-rigid structure from motion \ml[1]{(NRSfM)} problem of which the goal is to reconstruct deforming objects or scenes from their 2D trajectories. \ml[1]{NRSfM is in many ways different from 3D pose estimation, most importantly in that NRSfM is a 3D reconstruction problem based on geometric constraints while 3D pose estimation is more of a data-driven machine learning problem.}
\mlr{However, this paper has many novelties compared to []}{Compared to \cite{lee:consensus:cvpr16}, this work has many novel contributions}:
(i) \ml{I}t is the first work, to the best of our knowledge, applying the concept of \ml{multiple partial hypotheses} to 3D pose estimation, \ml[1]{and} furthermore, it shows the state-of-the-art performance on \mlt[1]{the }popular data sets, namely, Human3.6M \cite{Ionescu:h36m:pami14} and HumanEva \cite{sigal:humaneva:ijcv10} data sets.
(ii) Unlike \cite{lee:consensus:cvpr16}, the proposed scheme can be \ghr[2]{\ml[1]{trained}}{fine-tuned} in an end-to-end fashion, which \ml[1]{is beneficial for improving the overall} performance.
(iii) A domain conversion layer, which transforms heatmap representations to 2D coordinate representations, is proposed for an end-to-end \ghr[2]{training}{fine-tuning}.}



\ght[4]{The remainder of this paper is organized as follows.
\gh[1]{In Section \ref{sec:related}, we introduce \ght[4]{some }related works,} \gh[4]{and an overview of the proposed framework is described in Section \ref{sec:overview}.}
In Section \ref{sec:candidate}, the proposed joint group selection scheme and the partial 3D pose estimation method \ml{for the} selected joint group\ml{s} are introduced.
The proposed aggregation scheme is \ghr[4]{introduced}{explained} in Section \ref{sec:consensus}, \gh[1]{and the proposed end-to-end \ghr[2]{learning}{fine-tuning} scheme is introduced in Section \ref{sec:endtoend}.}
The experimental results are described in Section \ref{sec:exp}, and we conclude the paper in Section \ref{sec:conclusion}.}

\sh[1]{\section{Related Work}\label{sec:related}}
\gh[1]{In this section, we introduce \ght[3]{some }related works of single-image-based 3D pose estimation.}
They are following one of \mlt{the }two \mlr{main}{major} trends: (i) \ml{a} direct regression of 3D human pose\ml{s} from an image \cite{chen:synthesizing:3dv16, rogez:mocap:nips16, park:3dhumanpose:eccv16, pavlakos:coarse:arxiv16}, and (ii) \ml{a} two-step approach \ml{that first} estimates 2D joint coordinates from an image and \mlr{lifts the 2D pose to the 3D pose}{then infers the 3D pose from the 2D pose} \cite{zhou:sparseness:cvpr16, yasin:dualsource:cvpr16, chang:consistency:17,  chen:3dpose2dposepmatching:cvpr17, martinez:asimply:iccv17}.
The first approach has an advantage that the entire network could be learned in an end-to-end fashion, which might result in better performance.
Chen et al. \cite{chen:synthesizing:3dv16} \mlr{automatically synthesized the}{proposed a method to automatically synthesize} images with ground truth 3D poses to handle the issue of insufficient training samples.
However, the performance of the 3D pose estimator trained on the synthesized samples was poor on \mlt{the }real images, which required an additional domain adaptation process. 
Rogez et al. \cite{rogez:mocap:nips16} also synthesized \mlt{the }pairs of image\ml{s} and 3D pose\ml{s} based on 3D motion capture data.
Given the synthesized data, they clustered the 3D poses and formulated \mlt{the }3D human pose estimation \mlt{problem }as a 3D pose classification problem.
However, it is hard to guarantee \ml{that} the synthesized data follows the real-world data distribution.
Park et al. \cite{park:3dhumanpose:eccv16} proposed an end-to-end network which estimates both 2D pose and relative 3D joint coordinates with respect to multiple root joints directly from a single image.
At the test time, they averaged multiple relative 3D joint coordinates to infer the final 3D pose, which \ght[1]{is naive and }\ml{can be} sensitive to outliers\mlt{ because they just averaged the candidates}.
Pavlakos et al. \cite{pavlakos:coarse:arxiv16} proposed a different representation \ml{for} 3D poses other than 3D coordinates.
They utilized a voxelized 3D coordinate space for the new 3D representation, and estimated \ml{the} voxel-wise likelihood of each joint from an image.
However, the dimension of the voxel space is too high, which \ml{can} be a burden in \ml{the} training process.
Furthermore, voxel quantization lower\ml{s} the resolution of \mlt{the }3D space, which could worse\ml{n} the performance.

\mlr{In the second trend, they use the two-step approach which firstly estimates 2D pose from an image and lifts the 2D pose to the 3D pose.}{On the other hand, the second approach is a two-step approach that estimates the 2D pose first from an image and then reconstructs the 3D pose from the obtained 2D pose.}
Yasin et al. \cite{yasin:dualsource:cvpr16} and Chen et al. \cite{chen:3dpose2dposepmatching:cvpr17} estimated the 2D pose based on a pictorial structure model and a CNN-based model, respectively.
After that, \ml{the} 3D pose is retrieved based on \mlr{a ready-made 3D pose database using k-nearest neighbor with the estimated 2D pose}{the $k$-nearest neighbor samples of the estimated 2D pose in a ready-made 3D pose database}.
However, the framework of \cite{yasin:dualsource:cvpr16} is \mlr{iterative which}{based on an iterative procedure that} is hard to guarantee the convergence and \cite{chen:3dpose2dposepmatching:cvpr17} synthesized a 3D pose database which is hard to ensure that it follows the real-world 3D pose distribution.
Chang et al. \cite{chang:consistency:17} proposed a conditional\ml{-}random\ml{-}field\ml{-}based model over 2D poses.
In the model, the 3D pose is estimated as a byproduct of the inference process.
The unary term was defined based on the heatmap of a CNN\ml{-}based 2D pose estimator, and the prior term was defined based on the consistency of the estimated 2D pose and the reprojected 3D pose.
However, the camera parameters are needed to measure the consistency, which limits \mlt{the }applicable data.
Martinez et al. \cite{martinez:asimply:iccv17} proposed a fully-connected\ml{-}layer\ml{-}based lifting network\ml{,} applying recent techniques such as residual connections, batch normalization, and \ml{a\ght[1]{(?)}} max-norm constraint.
The proposed model is quite simple yet shows superior performance.
Fang et al. \cite{fang:grammar:aaai18} utilized some human body dependencies and relations \mlr{to}{in\ght[1]{(?)}} the 3D human pose estimation.
However, all of the two-step approaches including \cite{martinez:asimply:iccv17, fang:grammar:aaai18} are difficult to \mlr{be trained}{train} in an end-to-end fashion, which prevents the potential of future performance improvement.

\gh[4]{\section{Overview of the propose algorithm}\label{sec:overview}
An overview of the proposed method is visualized in Figure \ref{fig:overview}.
In our framework, several joint groups are selected based on the proposed weighted sampling process.
The probability distribution is designed to put the joints that have implicit interdependent movements into the same group.
After that, each joint group is modeled separately to estimate each partial 3D pose.
The proposed joint group selection scheme and the partial 3D pose estimation method \ml{for the} selected joint group\ml{s} are introduced in Section \ref{sec:candidate}.
Finally, partial 3D poses of \ml{the} joint groups are aggregated so that the 3D pose of the whole body is estimated based on the proposed robust optimization formula, which is introduces in Section \ref{sec:consensus}.}

\section{Weak pose estimation}
\label{sec:candidate}
In this section, we will introduce the joint group selection scheme and the 3D pose estimator \mlr{design for each}{for a} selected joint group.
Before explaining the proposed schemes, we introduce some notational conventions.
The input RGB image is denoted as $I$, and the 2D human pose is represented by \shr[1]{\ml{a} $2n$-dimensional vector $\mb{x}$}{$\mb{x} \in \mathbb{R}^{2n}$}.
Here, $n$ is the \ml{total} number of \mlt{whole }body joints and $\mb{x}$ is the stack of 2D coordinates of all the joints.
Similarly, the 3D human pose is represented by \shr[1]{\ml{a} $3n$-dimensional vector $\mb{X}$}{$\mb{X} \in \mathbb{R}^{3n}$}, which is the stack of 3D coordinates of all the joints.
From the $n$ joints, we select some overlapping joint groups.
The $j$th joint group is represented as $g_j = \{g_{j1}, \cdots, g_{j n_g}\}$, where the elements of $g_{j}$ are the joint indexes \mlt{which are }included in the $j$th joint group and $n_g$ is the number of joints in a joint group.

\subsection{Joint group selection}
We design the joint group selection scheme \mlr[3]{with \ml[3]{an} expectation}{hoping} that the complexity of \ml{a} group pose estimator is lower than \ml{a} whole\ml{-}body pose estimator.
To \mlr{achieve this}{realize this expectation}, we put joints that are interdependent to \mlt{be included in }the same joint group\ml[1]{, and these groups are sampled based on training data}.
\gh[1]{Here, we \mlr[1]{facilitate}{make use of} the fact that most of the \ml[1]{training} data sets for 3D pose estimation are composed of video sequences, which allows us to use the trajectory information.
The interdependency is evaluated based on the similarities of trajectories among joints, where} the similarity between the $i$th joint and the $i'$th joint is evaluated as
\begin{equation}
	s_{ii'} \triangleq \sum_{k \in \mathbb{S}} \|\mb{X}_{ik}-\mb{X}_{i'k}\|^2,
\end{equation}
where $\mathbb{S}$ is a \ghr[1]{set}{sequence} of training samples, $\mb{X}_{ik}$ and $\mb{X}_{i'k}$ are the 3D coordinates of the $i$th and the $i'$th joint from the $k$th sample, respectively.
Based on the similarity measure, the elements of a joint group are sequentially sampled based on a weighted sampling process.
The weight of the $i$th joint for the $j$th joint group is defined as
\begin{equation}
    \begin{split}
    w_i & \triangleq \left\{ \begin{array}{ll}
    \textrm{exp}\left(-\frac{\lambda}{2n_g}\sum_{i'\in g_j}s_{ii'}\right), & \text{if } i \notin g_j,  \\
    0, & \text{otherwise,}
    \end{array}
    \right.\\
	 \end{split} 
     \label{eqn:weights}
\end{equation}
where $\lambda$ is a predefined parameter.
Here, the first element of the joint group is selected based on a uniform distribution, and the remaining elements are selected based on the weighted sampling process with the weights $w_i$.
The joint group selection process is iterated until all the joints are included in joint groups at least $m_g$ times.
Let $n_t$ be the total number of joint groups.
\ml{Note} that the selected joint groups are commonly used for all the \ghr[1]{images}{samples \ml[1]{in} a data set.}

\ght[1]{It could sound awkward to use joint trajectories in a single-image-based application.
However, most of the 3D human pose estimation data sets consist of video sequences due to the difficulties in obtaining ground truth 3D coordinates.
To collect a large number of samples, the samples are continuously captured in the studio equipped with motion capture system.
In addition to this, the proposed scheme only needs a sequence of joint trajectory, which shows the generality of the scheme.}

\begin{figure*}[t]
    \centering
    \includegraphics[height=3.7cm]{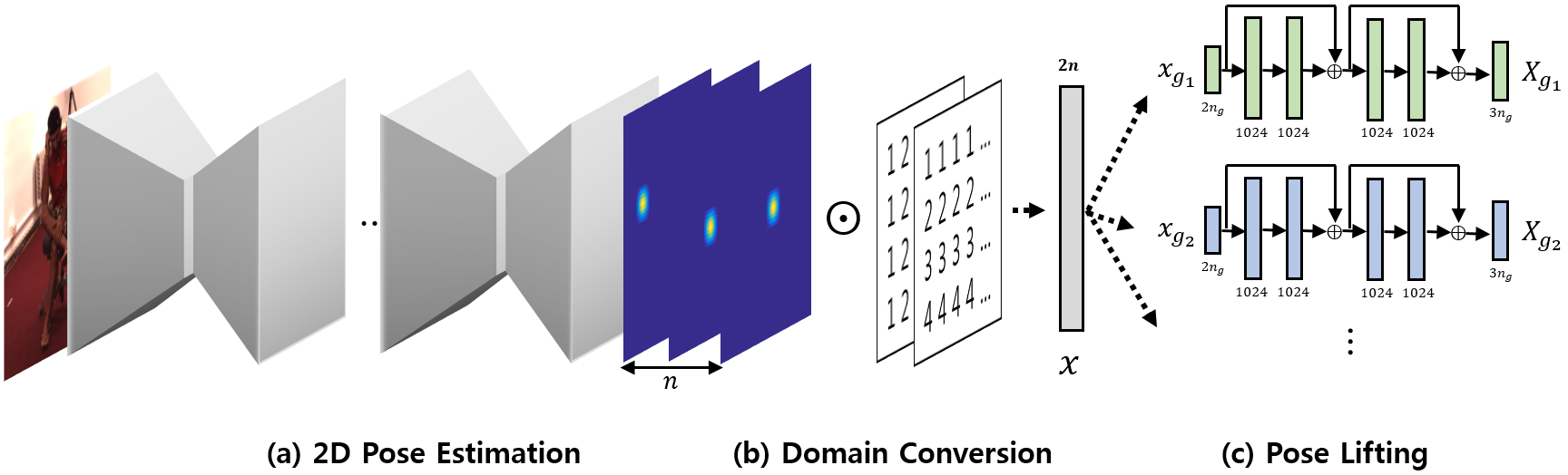}
    \caption{\ml{A} visualization of the 2D pose estimator network and the 3D pose lifting network. The 2D pose estimator consists of \mlt{concatenated }eight hourglass modules, and the output of the network is heatmap\ml{s} of joints. The 3D pose lifting network consists of fully-connected layers. The input of the 3D pose lifting network is 2D joint coordinates. \mlr{For}{To apply} an end-to-end \ghr[2]{training}{fine-tuning}, we need a differentiable domain conversion layer which is introduced in Section \ref{sec:endtoend}.}
    \label{fig:network_overview}
\end{figure*}

\subsection{Group pose estimation}
In this section, we explain \ml{the} 3D pose estimation scheme for the joint groups.
Here, we consider two \ghr[3]{kinds of the situations}{cases}\ml{,} following the general practice in the literature \cite{zhou:sparseness:cvpr16, martinez:asimply:iccv17}: (i) estimating 3D poses when the ground truth 2D poses are given \gh[1]{(\mlt[1]{we call it }``\ghr[3]{Situation}{Case} 1")}\ml{,} and (ii) estimating 3D poses when only RGB images are given without \ml{any} ground truth 2D poses \gh[1]{(\mlt[1]{we call it }``\ghr[3]{Situation}{Case} 2")}.\\\\
\textbf{\gh[1]{\ghr[3]{Situation}{Case} 1}\ml{.}}~
In this case, all we need to do is lifting the given 2D poses to the \ml{corresponding} 3D poses.
This task \ghr[1]{could}{can} be expressed as
\begin{equation}
	\mb{\hat{X}}_ {g_j} = f_j(\mb{x}_{g_j}),
	\label{eqn:liftingrel}
\end{equation}
where $\mb{\hat{X}}_{g_j}$ is the estimated 3D pose vector of the $j$th joint group, $\mb{x}_{g_j}$ is the ground truth 2D pose vector of the $j$th joint group, and $f_j$ is the 3D pose lifter for the $j$th joint group.
For this task, we could adopt any kinds of 3D lifter as $f_j$, though\mlt{,} we choose \ml{a} 3D lifter that is lightweight\ml{ed} and shows good performance.
In particular, the network size \ml{is} important to our approach \mlr{to handle the several joint groups effectively}{in respect of efficiency, because there are several joint groups to be lifted}.
With these considerations \ml{in mind}, we choose \cite{martinez:asimply:iccv17}\mlr{ for the pose lifter}{,} whose \mlr{network size}{complexity} is \mlr{not huge and}{relatively small even though it} shows the state-of-the-art performance\ml{, for \ghr[3]{the}{a} pose lifter}.

The chosen pose lifter \cite{martinez:asimply:iccv17} is designed based on a neural network.
A visualization of the lifting networks is shown in Figure \ref{fig:network_overview}\gh[1]{(c)}.
The input of the network is the $2n_g$-dimensional partial 2D pose vector \ml{of a selected joint group}.
The input is transformed to \ml{a} $1024$-dimensional vector \gh[1]{with a fully-connected layer}, which is fed into \ml{a} fully-connected network which consists of two cascaded blocks.
Each block has two consecutive fully-connected layers.
After each fully-connected layer, a batch normalization layer, a ReLU activation layer, and a dropout layer are followed.
The output of the fully-connected network is \ml{a} $1024$-dimensional vector, which is \ml[3]{again} transformed to \ml{a} $3n_g$-dimensional partial 3D pose vector \ml[3]{based on another fully-connected layer}.
For the detailed structure of the network, please refer to \cite{martinez:asimply:iccv17}.
\gh[1]{Here, we \mlt[3]{practically }found out that using a Leaky ReLU activation layer instead of the ReLU activation layer enhances the performance, therefore, we have changed the activation layers of the chosen 3D pose lifter \ml[3]{accordingly}.}

From (\ref{eqn:liftingrel}), we \mlt{could }derive a loss function for the 3D lifter as
\begin{equation}
	L_{\textrm{lifting}} =\sum_k \sum_j \|\mb{X}_{g_j}^k - f_j(\mb{x}_{g_k}^k)\|_F^2,
\end{equation}
where $\mb{X}_ {g_j}^k$ and $\mb{x}_ {g_j}^k$ are the ground truth 3D poses and the ground truth 2D poses of the $j$th joint group \ghr[2]{from}{in} the $k$th training sample, respectively.
Note here that each joint group has its own pose lifter.
\mlr{This process might be designed with a pose lifter, however,}{Alternatively, we might train a single pose lifter to handle all of the joint groups. However,} \mlr{we found out empirically that a single pose lifter could not give a reasonable solution for all the joint groups due to the increased modeling complexity.}{it is obvious that this approach is worse than using a different lifter for each group, as we have confirmed empirically \gh[3]{in Appendix}.}\\\\
\textbf{\gh[1]{\ghr[3]{Situation}{Case} 2}\ml{.}}~
In this \ml{case}, we use the two-step approach which first detects \ml{a} 2D pose $\mb{x}$ from the image $I$ and lifts \ml{the} 2D pose $\mb{x}$ to \ml{a} 3D pose $\mb{X}$.
The process of the \ml{latter} step is the same as that \ght[3]{in the case }of \ghr[1]{`given 2D poses'}{``\ghr[3]{Situation}{Case} 1''}.
Therefore, we only need \ml{the part} to estimate \ml{the} 2D pose from the input image.
A thing to note here is that \mlt{the }most of the high-performance 2D pose estimators output the results in \mlr{a heatmap representation}{heatmaps}.
\ml{Considering this, the} 2D pose estimation step \ml{can} be expressed as
\begin{equation}
	M(\mb{\hat{x}}) = h(I),
    \label{eqn:2dpartesti}
\end{equation}
where $\mb{\hat{x}}$ is the estimated 2D pose vector, $h$ is the 2D pose estimator, and $M$ is \ml{a} mapping function which converts 2D joint coordinates to \mlr{its heatmap representation}{their corresponding heatmaps}.

\mlr{Similarly, for the 2D pose estimator $h$,}{Similar to the 3D pose lifter,} we \ml{can} adopt any kinds of estimator \ml{for the 2D pose estimator $h$}.
\mlr{However, because}{Since} the accuracy of the 2D pose is very crucial for 3D lifting performance, we choose \ml{a} state-of-the-art 2D human pose estimator \cite{newell:stakced:eccv16}.
A visualization of the \ml{selected} 2D pose estimator is shown in Figure \ref{fig:network_overview}\gh[1]{(a)}.
It consists of \mlr{cascaded eight}{eight cascaded} hourglass modules, and each hourglass module consists of successive max-pooling layers and up-sampling layers.
Before each max-pooling \ml{or\ght[1]{(?)}} up-sampling layer, there is a residual module.
The output of the last hourglass module is fed into a $1$x$1$ convolution layer, which results in the heatmaps of the estimated 2D joint positions.
Here, we empirically found out that adding a sigmoid layer to the output of the estimator facilitate\ml{s} the training process.
\mlr{Hence, we used the modified version as the 2D pose estimator.}{Hence, we have added this modification to the estimator.}
The mapping function $M$ \ml{in} \cite{newell:stakced:eccv16} converts the 2D joint coordinates to \ghr[1]{\ml{a(?)}}{} $64\times 64$-size \ghr[1]{heatmap}{heatmaps} of which the value \ml{represents a} Gaussian distribution \ml{where the} mean is \ml{a} joint position and the variance is $3$. 
From (\ref{eqn:2dpartesti}), \mlt{we could get }a loss function for the 2D pose estimation \ml{can be given} as
\begin{equation}
	L_{\textrm{2d}}=\sum_k\sum_i\|M(\mb{x}_{i}^k) - h_i(I_k)\|_F^2,
\end{equation}
where $I_k$ is the RGB image \ml{of} the $k$th sample, $h_i$ is the heatmap of the $i$th joint that is estimated based on $h$, and $\mb{x}_i^k$ is the ground truth 2D coordinate\ml{s} of the $i$th joint \ml{in} the $k$th sample.

After \ml{estimating} the heatmap of each joint\mlt{ is estimated}, the \mlt{maximum }2D position \gh[2]{is obtained based on the proposed domain conversion layer which will be introduced in Section \ref{sec:endtoend}.} \gh[2]{The estimated 2D position from} each heatmap is concatenated to form \ghr[1]{$\mb{x}$}{$\mb{\hat{x}}$}.
\ml{For the 2D pose estimation process}, we use \ml{a} single \mlt{2D pose }estimator, i.e., it is not separately designed for each joint group, because \gh[2]{correlations among joints \ml[4]{are more} important in 2D pose estimation, \ml[4]{unlike in} the 3D \ml[4]{lifter}.}

\section{Consensus of joint groups}
\label{sec:consensus}
\mlr{Until now}{So far}, joint groups \mlr{are}{have been} selected, and \ml{the corresponding} partial 3D poses \mlr{of joint groups are}{have been} estimated.
\gh[1]{At the test time, partial 3D poses are aggregated to estimate the whole 3D pose based on the proposed robust optimization formula \mlt[3]{which will be }introduced in this section.}
\ght[1]{In this section, we introduce a scheme \ml{for} aggregating \mlr{joint group}{the partial} poses to estimate the \mlt{final }whole 3D pose.}
Before explaining the proposed scheme, we introduce some additional notations.
For a \ml[3]{3D pose vector} \ghr[1]{$3n$-dimensional vector $\mb{A}$}{$\mb{A} \in \mathbb{R}^{3n}$}, \ghr[1]{$\mb{A}|_{g_j}$}{$\mb{A}|_{g_j} \in \mathbb{R}^{3n_g}$} \ml[3]{indicates} the \ght[1]{$3n_g$-dimensional }subvector of $\mb{A}$, which consists of 3D coordinates of joints that are included in the set $g_j$, and \mlt{let }\ghr[1]{$\mb{A}'$}{$\mb{A}'\in\mathbb{R}^{n\times 3}$} \ml{is a} \ght[1]{$n\times 3$ }matrix whose rows are filled with the 3D coordinates of $\mb{A}$. 

\mlr{When designing the aggregation process, we should consider two issues}{There are two issues to consider when designing the aggregation process}, which are \ml{the} translation ambiguities between the estimations and the possibility of poor estimations in some joint groups.
The first issue could be resolved \ml{based on} the fact that there are overlapping joints between \ml{the} joint groups, i.e., we could reveal the translations with the constraints that the coordinates of the overlapping joints should be the same.
To deal with the second issue, we adopt \ml{the} median statistic that is robust to outliers.
It is well\ml{-}known that we \ml{can} obtain the median with \ml{an $l_1$}-norm minimization problem.
Keeping these in mind, we formulate the following problem:
\begin{equation}
	\mb{X} = \textrm{argmin}_{\mb{X}, \mb{t}_j}\sum_j \|\mb{X|}_{g_j}-\mb{\hat{X}}_{g_j}-\mb{1}\otimes\mb{t}_j\|_1,
    \label{eqn:l1concen}
\end{equation}
where $\mb{1}$ is the vector of ones, $\otimes$ is the Kronecker product, $\|\cdot\|_1$ is the \ml{$l_1$}-norm, and $\mb{t}_j$ is \ml{a} $3$-dimensional vector which represents the translation component of the $j$th joint group.
Note that this formulation does not \mlr[3]{respect the Euclidean structure when handling the outliers, i.e., it}{evaluate the error of a 3D point isotropically, because the $l_1$-norm} handles \ml[3]{each} coordinate independently.
\mlr{However, in this problem}{Instead}, we \ml{can} \mlr{take the Euclidean structure into account by using}{incorporate} the group sparsity \mlr{improve the performance}{to resolve this issue}.
\ml{Accordingly, the} formulation is modified as
\begin{equation}
	\mb{X}' = \textrm{argmin}_{\mb{X}',\mb{t}_j}\sum_j \|\mb{X'|}_{g_j}-\mb{X}'_{g_j}-\mb{1}\otimes\mb{t}_j^T\|_{2,1},
\end{equation}
where $\|\cdot\|_{2,1}$ is the \ml{$l_{2,1}$}-norm.
\ml{This} can \ml{also be} expressed as
\begin{equation}
	(\mb{X}', t) = \textrm{argmin}_{\mb{X}',\mb{t}}\left\|\begin{bmatrix} \mb{E}~~ & \mb{I}\otimes \mb{1}\end{bmatrix} \begin{bmatrix} \mb{X}' \\ \mb{t} \end{bmatrix}-\mb{F}\right\|_{2,1},
\end{equation}
where $\mb{I}$ is the identity matrix, and $\mb{E}$, $\mb{F}$, and $\mb{t}$ are defined as
\begin{equation}
	\begin{split}
    \mb{E} = \begin{bmatrix}
		\mb{E}_1, & \mb{E}_2, & \cdots 
	\end{bmatrix}^T,~~
    \mb{F} = \begin{bmatrix}
		\mb{X}'_{g_1}, & \mb{X}'_{g_2},  & \cdots
	\end{bmatrix}^T, ~~
	\mb{t} = \begin{bmatrix}
		\mb{t}_1, & \mb{t}_2, & \cdots
	\end{bmatrix}^T\ml{.}
    \end{split}
\end{equation}
\ml{Here,} $\mb{E}_j$ is \ml{an} $n \times n_g$\mlt{ size} matrix \ml{each of} whose columns \ml{is a} one-hot column vector\mlt{s} \mlr{which indicate the joint group membership}{that represents the index of each joint in the $j$th joint group}.

\ml{This} problem \ml{can} be solved with the alternating directional method of multipliers (ADMM) \cite{lin:ADMM:2010}, with an auxiliary variable $\mb{N}$.
The problem is modified as
\begin{equation}
	(\mb{G},\mb{N}) = \textrm{argmin}_{\mb{G},\mb{N}} \frac{\mu}{2} \|\mb{N}-\begin{bmatrix} \mb{E}~~ & \mb{I}\otimes \mb{1}\end{bmatrix}\mb{G}+\mb{F}\|_{F}^2+\|\mb{N}\|_{2,1}, 
\end{equation}
where $\mb{G}\triangleq \begin{bmatrix} \mb{X}', \mb{t} \end{bmatrix}^T$, and $\mu$ is a parameter.
\mlr{The problem is resolved}{The solution of this problem can be obtained} by solving $\mb{G}$ and $\mb{N}$ alternatively until convergence.
Note here that both $\mb{G}$ and $\mb{N}$ have closed-form solution\ml{s} \mlr{which need}{based on} a pseudo-inverse operation and a soft-thresholding operation, respectively.

\section{End-to-end \ml[3]{learning for} \gh[2]{fine-tuning}}\label{sec:endtoend}
We have introduced \ml{a} two-step algorithm for \mlt{the }3D human pose estimation.
However, \mlr{we could expect an improved performance when}{there is a chance to improve the overall performance if} we \ghr[2]{train}{fine-tune} the whole \mlr{network}{framework} in an end-to-end fashion.
In this section, we will introduce how \mlr{we do an end-to-end training}{this is possible}.
\\\\
\textbf{Consensus cost}\ml[1]{.}~
In the \gh[1]{proposed} two-step algorithm, \ghr[1]{\mlr{the aggregation procedure has done with another optimization process after the partial 3D poses of joint groups are estimated.}{the partial 3D poses have been aggregated based on a hand-designed optimization process that is independent with the training process.}}{the 3D lifting network of each joint group has been trained separately.}
However, we \ml{can\ght[1]{ instead}} add a loss of \mlt{the }aggregation \mlt{process }to the objective function of the 3D lifting network, \ml[3]{to improve the overall performance further}.
The aggregation cost \ml{is} defined as
\begin{equation}
	L_{\textrm{aggre}} = \sum_j \|\mb{X|}_{g_j}\mb{W}_j-\mb{\hat{X}}_{g_j}\|_F^2,
    \label{eqn:lconsen}
\end{equation}
where $\mb{W}_j\triangleq \mb{I}-\frac{\mb{w}_j\mb{w}_j^T}{n_g}$, and $\mb{w}_j$ is the $n$-dimensional vector of which the element is $1$ if it is included in the $j$th joint group and $0$ otherwise.
Here, the role of $\mb{W}_j$ is selecting the components of the $j$th joint group \ml{with} removing the translation component.
The exact \ml{translation of (\ref{eqn:l1concen}) to the} aggregation loss is \ml{to use} an \ghr[1]{\ml{$l_1$}}{$l_{2,1}$}-norm version of (\ref{eqn:lconsen}), but we empirically found out that the \ghr[1]{\ml{$l_1$}}{$l_{2,1}$-norm} unstabilize\ml{s} the backpropagation process\ml{. Hence}, we instead use the square of the Frobenius norm \ml{in} the aggregation loss.
\\\\
\textbf{Domain conversion}\ml[1]{.}~
The outputs of the 2D pose estimator are heatmaps of joints, and the inputs of the 3D lifter\ml[1]{s} are 2D joint coordinates.
Hence, for an end-to-end \ml[3]{training}, we need a \ght[3]{differentiable }\mlt[1]{conversion }layer which converts \mlt[1]{from }the heatmap\mlr[1]{ representation}{s} to \mlt[1]{the }2D coordinate\mlr[1]{ representation}{s}.
\gh[3]{An argmax layer could carry out this role in the forward pass, but it blocks \ml[4]{the} back-propagation \ml[4]{of gradients in the} training process since the argmax operation is not differentiable.}
\gh[3]{To resolve this,} we propose a novel \gh[3]{differentiable} \mlr[1]{linear}{domain conversion} layer\ght[3]{ which \mlr[1]{successfully doing this role}{can successfully perform this role}}.
Before proposing the conversion layer, we introduce \ght[2]{an }additional notation\gh[2]{s}.
Let $[a:b]$ be a $(b-a+1)$-dimensional vector of which the element\ml[1]{s} are \mlr[1]{gradually}{monotonically} increasing integers \mlt[1]{staring }from $a$ \mlr[1]{and ending at}{to} $b$\gh[2]{, and \mlr[3]{the normalized version of a heatmap $h(I)$ is denoted by $\tilde{h}(I)$}{let $\tilde{h}_i(I)$ be a normalized version of $h_i(I)$ so that the sum of all values becomes one}.}
\mlr[1]{We could convert from the heatmap representation to the 2D coordinate representation}{The heatmap} of the $i$th joint \ghr[1]{from}{in} the $k$th sample \ml[1]{is converted to 2D coordinates} as
\begin{equation}
	x_i = \sum_p \sum_q \big[\tilde{h}_i(I_k)\odot ([1:l_x] \otimes \mb{1}^T)\big]_{(p,q)}, ~~y_i = \sum_p \sum_q \big[\tilde{h}_i(I_k)\odot (\mb{1} \otimes [1:l_y]^T)\big]_{(p,q)},
\end{equation}
where $l_x$ and $l_y$ are the width and height of the heatmap\ml[1]{, respectively}, $\odot$ is the Hadamard product\ght[1]{ operation}, $\otimes$ is the Kronecker product\ght[1]{ operation}, \ml[1]{and} \ghr[1]{$p$ and $q$ are the indexes of the $x$-coordinate and the $y$-coordinate, respectively}{$\big[\mb{A}\big]_{(p,q)}$ is \mlt[3]{the operation extracting }the $(p,q)$-th element \ml[3]{of} $\mb{A}$}.
Note that the proposed layer only consists of linear operation\ml[1]{s}, which \ml[1]{are} differentiable.
\gh[3]{\ml[4]{A possible downside of this approach is that the outputs of the proposed layer may be different from those of an argmax layer, i.e.,} there might be a bias between the \ml[4]{two}. However, we empirically found out that there is \ml[4]{only} a slight difference, \ml[4]{as shown} in Appendix.}\\\\
\textbf{\ghr[2]{Learning}{Fine-tuning} procedure}\ml[1]{.}~
We empirically found out that training
the whole framework from the scratch results in a bad local optimum with poor performance.
Hence, we pre-train the 2D pose estimator and the 3D lifter based on the loss function\mlr[1]{ of}{s} $L_{\textrm{2d}}$ and $L_{\textrm{lifting}}+L_{\textrm{aggre}}$, respectively.
After that, we fine-tune the whole network in an end-to-end fashion with the following loss function:
\begin{equation}
	L_{\textrm{e2e}} = \alpha L_{\textrm{2d}} + L_{\textrm{lifting}} + L_{\textrm{aggre}},
\end{equation}
where $\alpha$ is a weighting parameter.

\begin{table*}[t]
\caption{Performance comparison result\ml[2]{s} on \ml[2]{the} Human3.6M data set in \mlt[2]{the case of }\gh[1]{``\ghr[3]{Situation}{Case} 1''} \ml[2]{with} ``Protocol 1.'' \gh[1]{Here, $n_g$ and $n_t$ are the number of joints included in a joint group and the total number of joint groups, respectively.}}
\centering
\resizebox{\textwidth}{!}{%
\begin{tabular}{l|rrrrrrrrrrrrrrr|r} \toprule
Protocol 1 & Direct. & Discuss & Eating & Greet & Phone & Photo & Pose & Purch. & Sitting & SittingD & Smoke & Wait & WalkD & Walk & WalkT & Avg \\ \toprule
Martinez et al. \cite{martinez:asimply:iccv17}         & 37.7 & 44.4 & 40.3 & 42.1 & 48.2 & 54.9 & 44.4 & 42.1 & 54.6 & 58.0 & 45.1 & 46.4 & 47.6 & 36.4 & 40.4 & 45.5 \\ \hline
Ours ($n_g=8$)  & 36.0 & 41.0 & 39.2 & 41.3 & 42.7 & 60.3 & 42.2 & 41.9 & 56.2 & 49.9 & 41.9 & 42.2 & 44.6 & 40.2 & 33.7 & 43.4 \\
Ours ($n_g=10$) & \bf{33.9} & \bf{40.5} & \bf{34.8} & \bf{38.6} & \bf{40.4} & \bf{50.2} & \bf{41.0} & \bf{39.8} & \bf{55.4} & 48.8 & \bf{39.4} & \bf{40.4} & \bf{42.3} & 36.5 & \bf{31.9} & \bf{40.9} \\
Ours ($n_g=13$) & 34.5 & 41.7 & 35.4 & 39.4 & 41.2 & 50.9 & 41.4 & 40.9 & 56.0 & \bf{47.8} & 40.0 & 41.1 & 43.7 & 35.6 & 32.6 & 41.5 \\
Ours ($n_g=15$) & 35.3 & 42.6 & 37.1 & 39.9 & 43.3 & 52.1 & 42.8 & 40.5 & 55.9 & 50.5 & 41.4 & 41.9 & 44.3 & 36.0 & 32.4 & 42.6 \\
Ours ($n_g=17$) & 35.0 & 42.4 & 37.0 & 39.7 & 43.7 & 52.7 & 42.2 & \bf{39.8} & 56.3 & 51.1 & 41.7 & 41.4 & 45.1 & \bf{35.0} & 32.1 & 42.6 \\
Ours ($n_g=17$, $n_t=1$) & 36.0 & 43.3 & 37.4 & 40.3 & 44.2 & 53.4 & 42.7 & 40.7 & 56.4 & 52.1 & 42.5 & 42.4 & 45.5 & 35.5 & 32.6 & 43.3 \\
\bottomrule
\end{tabular}%
}
\label{tab:given2d}
\end{table*}

\begin{table*}[t]
\caption{Performance comparison results on \ml[2]{the} Human3.6M data set in \mlt[2]{the case of }\gh[1]{``\ghr[3]{Situation}{Case} 2''} \ml[2]{with} ``Protocol 1.''}
\centering
\resizebox{\textwidth}{!}{%
\begin{tabular}{l|rrrrrrrrrrrrrrr|r} \toprule
Protocol 1 & Direct. & Discuss & Eating & Greet & Phone & Photo & Pose & Purch. & Sitting & SittingD & Smoke & Wait & WalkD & Walk & WalkT & Avg \\ \toprule
LinKDE \cite{Ionescu:h36m:pami14}  & 132.7 & 183.6 & 132.3 & 164.4 & 162.1 & 205.9 & 150.6 & 171.3 & 151.6 & 243.0 & 162.1 & 170.7 & 177.1 & 96.6 & 127.9 & 162.1 \\ 
Li et al \cite{li:maximum:iccv15} & - & 136.9 & 96.9 & 124.7 & - & 128.7 & - & - & - & - & - & - & 132.2 & 70.0 & - & - \\ 
Tekin et al \cite{tekin:direct:cvpr16} & 102.4 & 147.2 & 88.8 & 125.3 & 118.0 & 182.7 & 112.4 & 129.2 & 138.9 & 224.9 & 118.4 & 138.8 & 126.3 & 55.1 & 65.8 & 125.0 \\ 
Zhou et al \cite{zhou:sparseness:cvpr16} & 87.4 & 109.3 & 87.1 & 103.2 & 116.2 & 143.3 & 106.9 & 99.8 & 124.5 & 199.2 & 107.4 & 118.1 & 114.2 & 79.4 & 97.7 & 113.0 \\ 
Tekin et al \cite{tekin:structured:bmvc16} & - & 129.1 & 91.4 & 121.7 & - & 162.2 & - & - & - & - & - & - & 130.5 & 65.8 & - & - \\ 
Ghezelghieh et al \cite{ghezelghieh:learning:3dv16} & 80.3 & 80.4 & 78.1 & 89.7 & - & - & - & - & - & - & - & - & - & 95.1 & 82.2 & - \\ 
Du et al \cite{du:marker:eccv16} & 85.1 & 112.7 & 104.9 & 122.1 & 139.1 & 135.9 & 105.9 & 166.2 & 117.5 & 226.9 & 120.0 & 117.7 & 137.4 & 99.3 & 106.5 & 126.5 \\ 
Park et al \cite{park:3dhumanpose:eccv16} & 100.3 & 116.2 & 90.0 & 116.5 & 115.3 & 149.5 & 117.6 & 106.9 & 137.2 & 190.8 & 105.8 & 125.1 & 131.9 & 62.6 & 96.2 & 117.3 \\ 
Zhou et al \cite{zhou:kinematic:eccv16} & 91.8 & 102.4 & 96.7 & 98.8 & 113.4 & 125.2 & 90.0 & 93.8 & 132.2 & 159.0 & 107.0 & 94.4 & 126.0 & 79.0 & 99.0 & 107.3 \\ 
Pavlakos et al \cite{pavlakos:coarse:arxiv16} & 67.4 & 71.9 & 66.7 & 69.1 & 72.0 & 77.0 & 65.0 & 68.3 & 83.7 & 96.5 & 71.7 & 65.8 & 74.9 & 59.1 & 63.2 & 71.9 \\ 
Martinez et al \cite{martinez:asimply:iccv17} &  51.8 &  56.2  &  58.1 & 59.0 & 69.5 & 78.4 & 55.2 & 58.1  & 74.0 & 94.6 & 62.3 & 59.1  &  65.1  & 49.5 & 52.4  & 62.9  \\
Fang et al \cite{fang:grammar:aaai18} &  50.1 &  54.3  &  57.0 & 57.1 & 66.6 & \bf{73.3} & 53.4 & 55.7  & 72.8 & \bf{88.6} & 60.3 & 57.7  &  62.7  & 47.5 & 50.6  & 60.4 \\\hline
Ours & \bf{48.4} & \bf{52.9} & \bf{55.2} & \bf{53.8} & \bf{62.8} & \bf{73.3} & \bf{52.3} & \bf{52.2} & \bf{71.0} & 89.9 & \bf{58.2} & \bf{53.6} & \bf{61.0} & \bf{43.2} & \bf{50.0} & \bf{58.8} \\
\bottomrule
\end{tabular}%
}
\label{tab:givenimage}
\end{table*}

\section{Experimental results}
\label{sec:exp}

We have evaluated the proposed \mlr[2]{approach}{scheme} quantitatively and qualitatively on several data sets.
For the quantitative experiments, we applied the proposed scheme on popular benchmark data sets, namely \ml[2]{the} Human3.6M \cite{Ionescu:h36m:pami14} data set and \ml[2]{the} HumanEva-I \cite{sigal:humaneva:ijcv10} data set.
We also applied the proposed scheme on \ml[2]{the} MPII data set \cite{andriluka:mpii:cvpr14} for the qualitative evaluation.

Human3.6M is the largest data set, to the best of our knowledge, that has synchronized RGB images and the corresponding 3D joint coordinates.
Intrinsic and extrinsic camera parameters are also provided so that we can obtain the corresponding 2D joint coordinates.
It consists of $15$ actions (e.q., direction, discussion, eating, etc.), and every action is performed by $7$ actors.
Each demonstration is captured in $4$ different angles simultaneously. 
Following the standard practices in the literature \cite{martinez:asimply:iccv17, fang:grammar:aaai18}, the demonstrations of subject\ml[2]{s} $1$, $5$, $6$, $7$, and $8$ were used as the training set, and the demonstrations of subject\ml[2]{s} $9$ and $11$ were used as the test set.

HumanEva-I is a smaller data set compared to the Human3.6M data set.
It also has synchronized RGB images with the corresponding 3D joint coordinates and 2D joint coordinates.
Following the practices in \cite{yasin:dualsource:cvpr16, pavlakos:coarse:arxiv16, martinez:asimply:iccv17}, we evaluated on all subjects\mlr[2]{ and in each action separately}{, separately in each action\ght[2]{(?)}}.

MPII is \ml[2]{a} popular benchmark data set for 2D pose estimation, which has RGB images taken \ml[2]{``}in the wild\ml[2]{''} and the corresponding manually\ml[2]{-}annotated ground truth 2D joint coordinates.
It has no ground truth 3D poses.

We have compared the performance of the proposed scheme based on the Euclidean distance between the ground truth 3D coordinate and the inferred 3D coordinate\ml[2]{s} after an alignment.
The final performance is the average distance of all joints and test samples. 
In the literature \cite{martinez:asimply:iccv17, fang:grammar:aaai18}, two types of alignment methods have been used, therefore, we report the performance in both the cases.
In the first case, the average distances are measured after the root joint alignments between the ground truth 3D poses and \ml[2]{the} inferred 3D poses, and we call this \ml[2]{``}Protocol 1\ml[2]{''}.
In the second case, the average distances are measured after a rigid alignment, and we call this \ml[2]{``}Protocol 2\ml[2]{''}.
For all the experiments, we used the following parameter setting unless we notice: $m_g=10$, $\lambda = 10$, $\alpha = 100$, $n_g=n-1$, and we used the first sequence of the training set as $\mathbb{S}$.

\begin{table*}[t]
\caption{Performance comparison results on \ml[2]{the} Human3.6M data set in \mlt[2]{the case of }\gh[1]{``\ghr[3]{Situation}{Case} 2''} \ml[2]{with} ``Protocol 2.''}
\centering
\resizebox{\textwidth}{!}{%
\begin{tabular}{l|rrrrrrrrrrrrrrr|r} \toprule
Protocol 2 & Direct. & Discuss & Eating & Greet & Phone & Photo & Pose & Purch. & Sitting & SittingD & Smoke & Wait & WalkD & Walk & WalkT & Avg \\ \toprule
Akhter et al. \cite{akhter:posecondition:cvpr15}& 199.2 & 177.6 & 161.8 & 197.8 & 176.2 & 186.5 & 195.4 & 167.3 & 160.7 & 173.7 & 177.8 & 181.9 & 176.2 & 198.6 & 192.7 & 181.1 \\ 
Ramakrishna et al. \cite{ramakrishna:2dlandmark:eccv12}& 137.4 & 149.3 & 141.6 & 154.3 & 157.7 & 158.9 & 141.8 & 158.1 & 168.6 & 175.6 & 160.4 & 161.7 & 150.0 & 174.8 & 150.2 & 157.3 \\ 
Zhou et al. \cite{zhou:sparse:tpami17}& 99.7 & 95.8 & 87.9 & 116.8 & 108.3 & 107.3 & 93.5 & 95.3 & 109.1 & 137.5 & 106.0 & 102.2 & 106.5 & 110.4 & 115.2 & 106.7 \\ 
Bogo et al. \cite{bogo:smpl:eccv16}& 62.0 & 60.2 & 67.8 & 76.5 & 92.1 & 77.0 & 73.0 & 75.3 & 100.3 & 137.3 & 83.4 & 77.3 & 86.8 & 79.7 & 87.7 & 82.3 \\ 
Moreno-Noguer \cite{moreno:3dhumanpose:cvpr17}& 66.1 & 61.7 & 84.5 & 73.7 & 65.2 & 67.2 & 60.9 & 67.3 & 103.5 & 74.6 & 92.6 & 69.6 & 71.5 & 78.0 & 73.2 & 74.0 \\ 
Pavlakos et al \cite{pavlakos:coarse:arxiv16} & - & - & - & - & - & - & - & - & - & - & - & - & - & - & - & 51.9 \\ 
Martinez et al \cite{martinez:asimply:iccv17} &  39.5 & 43.2  & 46.4 & 47.0 & 51.0 & 56.0 & 41.4 & 40.6 & 56.5 & 69.4 & 49.2 & 45.0 & 49.5 & 38.0 & 43.1  & 47.7  \\ 
Fang et al \cite{fang:grammar:aaai18} & \bf{38.2} & \bf{41.7} & \bf{43.7} & \bf{44.9} & 48.5 & \bf{55.3} & 40.2 & \bf{38.2} & \bf{54.5} & \bf{64.4} & 47.2 & 44.3 & 47.3 & 36.7 & 41.7 & \bf{45.7} \\ \hline
Ours & 39.6 & \bf{41.7} & 45.2 & 45.0 & \bf{46.3} & 55.8 & \bf{39.1} & 38.9 & 55.0 & 67.2 & \bf{45.9} & \bf{42.0} & \bf{47.0} & \bf{33.1} & \bf{40.5} & \bf{45.7} \\
\bottomrule
\end{tabular}%
}
\label{tab:givenimage2}
\end{table*}

\subsection{Implementation details}
We used \ml[2]{a} separate 2D pose estimator\mlt[2]{s} on each data set\mlt[2]{s}, namely, \ml[2]{the} Human3.6M and \ml[2]{the} HumanEva-I data sets.
Both \ml[2]{the} 2D pose estimators were pre-trained on the MPII data set for $100$ epochs using RMSProp \cite{tieleman2012lecture}.
We used \ml[2]{an} exponential\ml[2]{ly-}decaying learning rate with \ml[2]{a} starting learning rate of $2.5\times 10^{-4}$.
After that, we fine-tuned the estimators for $20$ epochs using RMSProp with \ml[2]{the same} starting learning rate\mlt[2]{ of $2.5\times 10^{-4}$} which was reduced by the factor of $5$ after $10$ epochs.
In each training process, we used the batch size of $64$.
Here, in the fine-tuning process on Human3.6M, we uniformly sub-sampled the data set with the factor of $50$.

We also trained 3D pose lifters separately on each data set.
The 3D pose lifter\ml[2]{s were also} trained based on RMSProp with \ml[2]{an} exponential\ml[2]{ly-}decaying learning rate with \ml[2]{a} starting learning rate of $0.001$.
On \ml[2]{the} Human3.6M data set, the 3D pose lifter was trained for $100$ epochs with the batch size of $1024$.
On the other hand, because the number of training samples on HumanEva-I is smaller than $1024$, we used \ml[2]{a} smaller batch size of $64$, which was trained for $500$ epochs.
Here, the 3D pose lifter network was trained with the estimation results \ml[2]{from} each 2D pose estimator \mlt[2]{were used }as input.

The whole framework was fine-tuned for $5$ epochs using RMSProp, with \ml[2]{the} batch size of $128$, in \mlr[2]{the proposed}{an} end-to-end fashion \ml[2]{as proposed in Section \ref{sec:endtoend}}.
We used exponential\ml[2]{ly-}decaying learning rates of $1.25\times 10^{-4}$ (for the 2D pose estimator) and $0.0005$ (for the 3D pose lifter).
It is well-known that the first part of a network captures universal features like edges.
Therefore, we fixed the parameters of the first five hourglass modules.
\gh[2]{In the fine-tuning process on Human3.6M, we uniformly sub-sampled the data set with the factor of $50$ \gh[3]{\ml[4]{due to} the large number of training samples}.}
\mlr[2]{On the other hand,}{\ml[3]{Another} thing to note \mlt[3]{here }is that, for HumanEva-I, this end-to-end fine-tuning was not helpful, because} the HumanEva-I data set has too small number of training samples, resulting in \ml[2]{an} overfitting of the whole framework.

\begin{table}[t]
\caption{Performance on \ml[2]{the} HumanEva-I data set in \mlt[2]{the case of }\gh[1]{``\ghr[3]{Situation}{Case} 2''} \ml[2]{with} ``Protocol 2.''}
\centering
\resizebox{0.63\textwidth}{!} {
\begin{tabular}{l|ccc|ccc|c}
          \toprule
Protocol 2  &  \multicolumn{3}{c|}{Walking} & \multicolumn{3}{c|}{Jogging} & \multirow{2}{*}{Avg} \\ 
           & S1 & S2 & S3 & S1 & S2 & S3 &  \\ \toprule
          Radwan et al. \cite{radwan:monocular:iccv13} & 75.1 & 99.8 & 93.8 & 79.2 & 89.8 & 99.4 & 89.5 \\
          Wang et al. \cite{wang:robust:cvpr14}& 71.9 & 75.7 & 85.3 & 62.6 & 77.7 & 54.4 & 71.3 \\
          Simo-Serra et al. \cite{simo:joint:cvpr13}& 65.1 & 48.6 & 73.5 & 74.2 & 46.6 & 32.2 & 56.7 \\
          Bo et al. \cite{bo:twin:ijcv10}& 46.4 & 30.3 & 64.9 & 64.5 & 48.0 & 38.2 & 48.7 \\
          Kostrikov et al. \cite{kostrikov:depth:bmvc14}& 44.0 & 30.9 & 41.7 & 57.2 & 35.0 & 33.3 & 40.3 \\
          Yasin et al. \cite{yasin:dualsource:cvpr16}& 35.8 & 32.4 & 41.6 & 46.6 & 41.4 & 35.4 & 38.9 \\
          Moreno-Noguer et al. \cite{moreno:3dhumanpose:cvpr17}& 19.7 & \bf{13.0} & \bf{24.9} & 39.7 & 20.0 & 21.0 & 26.9 \\
          Pavlakos et al. \cite{pavlakos:coarse:arxiv16}& 22.1 & 21.9 & 29.0 & 29.8 & 23.6 & 26.0 & 25.5 \\
          Martinez et al. \cite{martinez:asimply:iccv17}& 19.7 & 17.4 & 46.8 & \bf{26.9} & 18.2 & 18.6 & 24.6 \\
          Fang et al \cite{fang:grammar:aaai18} & 19.4 & 16.8 & 37.4 & 30.4 & \bf{17.6} & \bf{16.3} & 22.9 \\\hline
          Ours & \bf{18.1} & 15.6 & 31.7 & 38.2 & 18.6 & 17.9 & \bf{22.5} \\ \bottomrule
          \end{tabular}%
        \label{tab:perfhumaneva}}
\end{table}

\begin{table}[t]
\caption{Ablation experiment\ml[2]{s} on different component\ml[2]{s} in our framework. It was performed on the Human3.6M data set \gh[3]{in ``Case 2''} with ``Protocol 1''.}
\centering
\resizebox{0.86\textwidth}{!} {
        \begin{tabular}{l|c|c|c|c|c|c} \toprule
          Variant  & Ours & w/o $L_{\textrm{e2e}}$ & w/o $l_{2,1}$&  w/o $L_{\textrm{aggre}}$  & \gh[4]{Random Selection} & \gh[4]{Random Selection, w/o $L_{\textrm{e2e}}$} \\ \toprule
          Error (mm) & 58.8 & 60.1 & 58.9 & 59.4 & 62.3 & 64.8 \\ \hline
          $\triangle$ &- & 1.3 & 0.1 & 0.6 & 3.5 & 6.0 \\  
          \bottomrule
          \end{tabular}%
        \label{tab:ablation}
        }
\end{table}

\subsection{Quantitative evaluation}
\textbf{\gh[1]{\ghr[3]{Situation}{Case} 1}\ml[2]{.}}~
We compared the performance of the proposed method to \ml[2]{that of} \cite{martinez:asimply:iccv17} \ml[2]{on Human3.6M data set,} in the \ghr[3]{situation}{case} that the ground truth 2D poses are given\mlt[2]{ on Human3.6M data set}.
In this experiment\mlt[2]{s}, we used \ml[2]{``}Protocol 1\ml[2]{''} for the evaluation, and the results are summarized in Table \ref{tab:given2d}.
\ml[2]{Here, we report} the performance\mlt[2]{s} with various values of $n_g$.
In all the cases, the proposed scheme shows \mlt[2]{the }superior performance to \cite{martinez:asimply:iccv17}.
In the result, we can \mlr[2]{demonstrate}{confirm} the effectiveness of the proposed \ghr[2]{part-based}{multiple-partial-hypothesis-based} approach.
As $n_g$ decreases from $17$ (the total number of joints) to $10$, the \mlr[2]{accuracies}{errors} are monotonically \mlr[2]{increase}{decreased}.
This result shows that modeling \ml[2]{a} partial joint group\mlt[2]{s} has \ml[2]{a} lower complexity compare\ml[2]{d} to modeling full joint\ml[2]{s}\mlt[2]{, resulting in the better performance}.
On the other hand, in the case of $n_g=17$, i.e., each 3D pose lifter estimates full 3D poses, we can demonstrate the effectiveness of the multiple-hypothesis-based model.
Even in the case that each 3D pose lifter estimate\ml[2]{s} full 3D pose, their aggregation has better performance.
\gh[2]{Finally, in the case of $n_g=17$ and $n_t=1$, \ml[4]{where only a single ``weak'' 3D lifter is used without combining multiple hypotheses, the proposed method is better than \cite{martinez:asimply:iccv17}. In this case, since the weak 3D lifter is a modified version of \cite{martinez:asimply:iccv17}, the only difference between the two is the modifications we made. Since our version gives better performance, this justifies the use of such modifications.}
\mlt[4]{we can \ml[4]{confirm that} the effectiveness of the modifications we applied to \ghr[3]{the 3D pose lifter}{the chosen 3D pose lifter [], i.e., it has lower error than that of []}.}}\\\\
\textbf{\gh[1]{\ghr[3]{Situation}{Case} 2}\ml[2]{.}}~
We compared the performance of the proposed \ghr[3]{method}{scheme} with various methods \cite{Ionescu:h36m:pami14, li:maximum:iccv15, tekin:direct:cvpr16, zhou:sparseness:cvpr16, tekin:structured:bmvc16, ghezelghieh:learning:3dv16, du:marker:eccv16, park:3dhumanpose:eccv16, zhou:kinematic:eccv16, pavlakos:coarse:arxiv16, martinez:asimply:iccv17, fang:grammar:aaai18, akhter:posecondition:cvpr15, ramakrishna:2dlandmark:eccv12, zhou:sparse:tpami17, bogo:smpl:eccv16, moreno:3dhumanpose:cvpr17} in the \ghr[3]{situation}{case} that only RGB images are given without any ground truth 2D poses on \ml[2]{the} Human3.6M data set.
The comparison results based on \ml[2]{``}Protocol 1\ml[2]{''} and \ml[2]{``}Protocol 2\ml[2]{''} are shown in Table \ref{tab:givenimage} and Table \ref{tab:givenimage2}, respectively.
In the case of \ml[2]{``}Protocol 1\ml[2]{''}, the proposed method shows the state-of-the-art performance.
On the other hand, \ml[2]{for} \ml[2]{``}Protocol 2\ml[2]{''}, the proposed scheme shows \ml[2]{almost} the same performance with \cite{fang:grammar:aaai18} \ml[2]{on average}.
However, \mlr[2]{noting that}{since} \ml[2]{``}Protocol 2\ml[2]{''} needs \ml[2]{a} rigid alignment\mlt[2]{s}, we \mlt[2]{can }claim that the proposed method has better performance.
We also compared the performance of the proposed method on \ml[2]{the} HumanEva-I data set with various methods \cite{radwan:monocular:iccv13, wang:robust:cvpr14, simo:joint:cvpr13, bo:twin:ijcv10, kostrikov:depth:bmvc14, yasin:dualsource:cvpr16, moreno:3dhumanpose:cvpr17, pavlakos:coarse:arxiv16, martinez:asimply:iccv17, fang:grammar:aaai18}.
The result is summarized in Table \ref{tab:perfhumaneva}.
The proposed method shows the best average performance, which demonstrate\ml[2]{s} the effectiveness of the proposed framework.

\begin{figure*}[t]
    \centering
    \includegraphics[height=4.0cm]{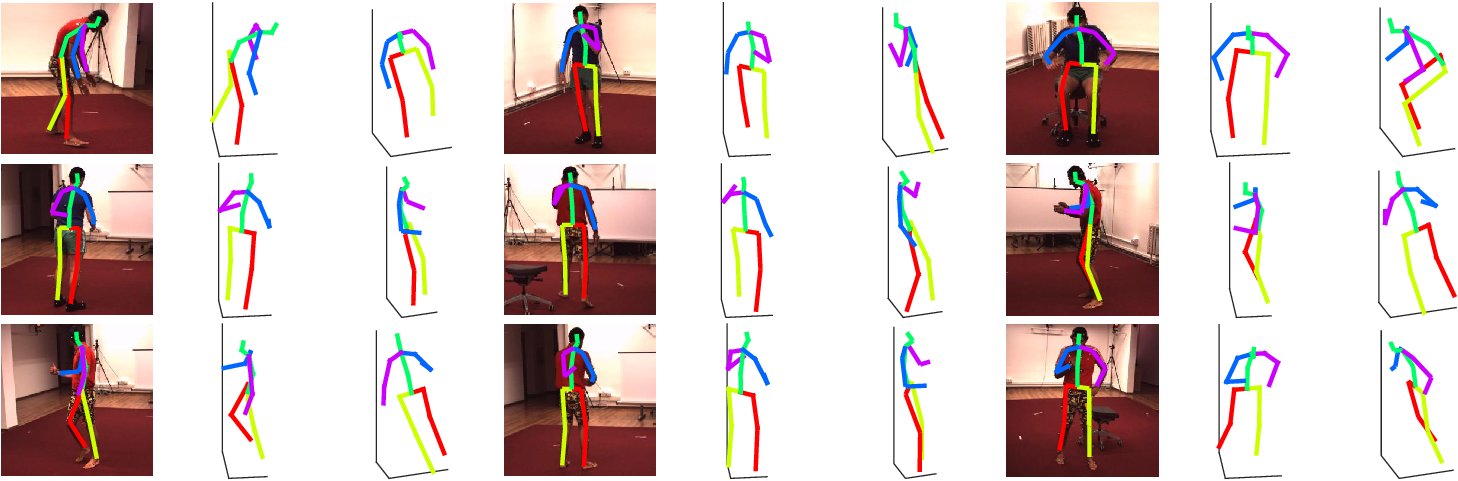}
    \caption{3D estimation results of some examples \ml[2]{in} the test set of \ml[2]{the} Human36M data set. Left: \ml[2]{I}nput RGB image\ml[2]{s} and \ml[2]{the} corresponding 2D pose estimation\ml[2]{s}. \gh[3]{Middle, }Right: 3D pose estimation result\ml[2]{s} \gh[3]{in two different views}.}
    \label{fig:example_human36m}
\end{figure*}

\begin{figure*}[t]
    \centering
    \includegraphics[height=5.5cm]{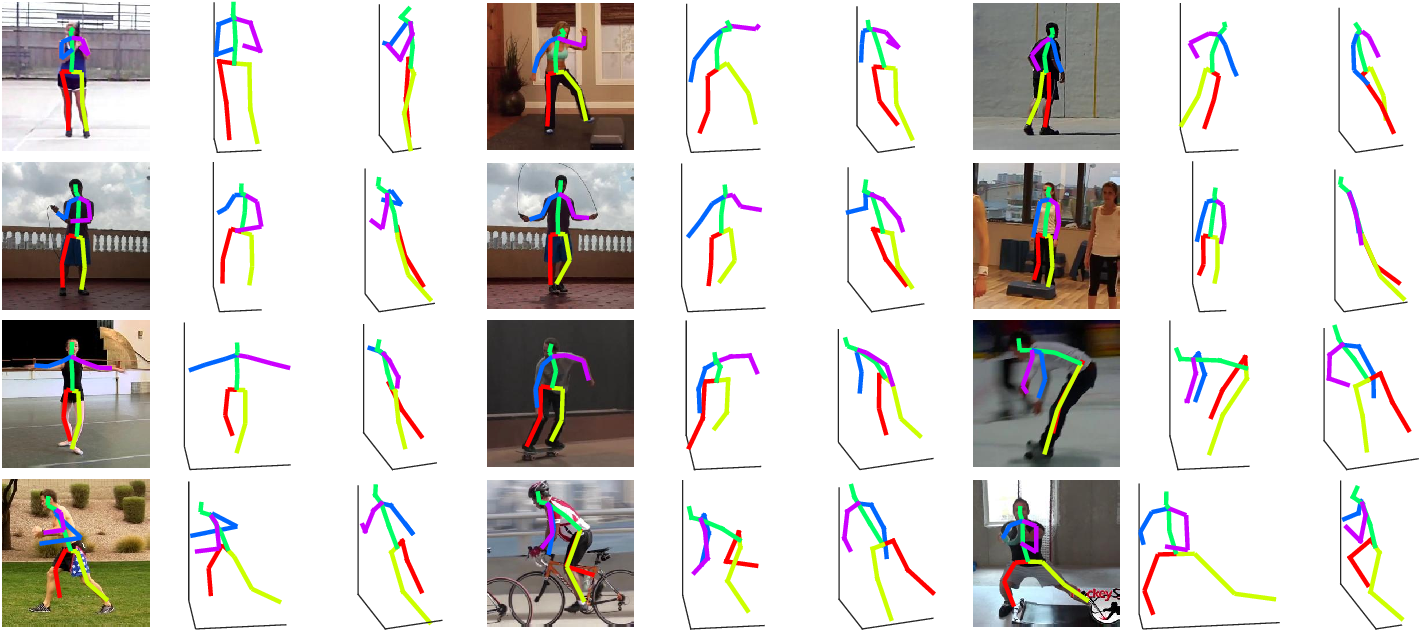}
    \caption{Some examples \ml[2]{in} the test set of \ml[2]{the} MPII data set. Left: \ml[2]{I}nput RGB image\ml[2]{s} and \ml[2]{the} corresponding 2D pose estimation\ml[2]{s}. \gh[3]{Middle, }Right: 3D pose estimation result\ml[2]{s} \gh[3]{in two different views}.}
    \label{fig:example_mpii}
\end{figure*}

\subsection{Ablation experiments}
We performed some ablation experiments on the Human3.6M data set with \ml[2]{``}Protocol 1\ml[2]{''}.
The performance \ml[2]{was} evaluated with removing some components of the proposed framework.
The results are summarized in Table \ref{tab:ablation}.
Removing the end-to-end fine-tuning process \ml[2]{increased} the error \mlt[2]{was increased }by 1.3mm.
The error was increased by 0.1mm when we \ml[2]{performed} the aggregation process with \ml[2]{the} $l_1$-norm objective function, and the error was increase\ml[2]{d} by 0.6mm when the aggregation loss $L_{\textrm{aggre}}$ was not included in the loss function.
\mlr[2]{On the other hand}{In the mean time}, when we randomly select\ml[2]{ed} joint group\ml[2]{s} without using the proposed selecting \ml[2]{scheme}, \gh[4]{the error was increased by 3.5mm, and removing the end-to-end fine-tuning process further increased the error by 2.5mm.}\ght[4]{\gh[3]{and \mlt[4]{we }removed the end-to-end fine-tuning process, the error was increased by 6.0mm, and the error was decreased by 2.5mm with the end-to-end fine-tuning process.}}

\subsection{Qualitative evaluation}
We report some qualitative results on \ml[2]{the} Human3.6M and \ml[2]{the} MPII data sets.
Some results on Human3.6M are visualized in Figure \ref{fig:example_human36m}, and \ml[2]{those} on MPII are visualized in Figure \ref{fig:example_mpii}.
Note that \ml[2]{the} MPII data set has no ground truth 3D poses, hence, for \ml[2]{this} data set, we \mlr[2]{applied images from MPII to}{used} the proposed network trained on \ml[2]{the} Human3.6M data set.
Unlike \ml[2]{the} images of Human3.6M, those of MPII are taken \ml[2]{``}in the wild\ml[2]{.''}
Although the proposed scheme was not trained on the wild images, we can see that the estimation results are reasonable.
\gh[3]{More results are available in Appendix.}


\section{Conclusion}
\label{sec:conclusion}
In this paper, we \mlr[2]{handled}{dealt with} the problem of 3D human pose estimation from a single image.
Single-image-based 3D human pose estimation is \ml[2]{a} very tough problem \ml[2]{due to} many reasons such as self-occlusions, wild appearance changes, and inherent ambiguit\ml[2]{ies} of 3D estimation from a 2D cue.
\mlt[2]{However, }\gh[2]{Most of the }\ml[2]{c}onventional methods have handled the problem with a single complex estimator, which have become requiring increasingly complex estimators to enhance the performance.
In this paper, we proposed a multiple-\ml[2]{partial\ght[2]{(?)}-}hypothesis-based framework for the problem.
We selected joint groups from the \mlr[2]{human joint model}{data} based on the proposed sampling scheme, and estimated partial 3D poses of \ght[2]{each }joint group\ml[2]{s} separately\ml[2]{. These were later} aggregated to obtain the final full 3D pose\mlt[2]{s} using the proposed robust optimization formula.
The proposed method is \ghr[2]{trained}{fine-tuned} in an end-to-end fashion, resulting in better performance.
In the experiments, the proposed framework shows the state-of-the-art performance on the popular benchmark data sets.
The proposed framework \mlr[2]{would}{can} be successfully adopted to \ml[2]{a} more general problem like multi-person 3D pose estimation based on a properly\ml[2]{-}designed joint\ml[2]{-}group selection scheme, which is left as a future work.

\bibliographystyle{splncs}
\bibliography{ref}

\setcounter{section}{0}
\renewcommand\thesection{\Alph{section}}
\renewcommand\thesubsection{\thesection.\arabic{subsection}}

\section{Appendix}
\subsection{\ml{A s}ingle 3D pose lifter for \ml{all} joint groups}\label{sec:singlelifter}

We introduce the performance of the proposed scheme in case we train a single pose lifter to handle all \sht[1]{of the }joint groups.
We have compared the results of training separate 3D pose lifter\ml{s} to those of training a single pose lifter on the Human3.6M data set \cite{Ionescu:h36m:pami14} in ``Case 1'' with ``Protocol 1''.
The results are summarized in Table \ref{tab:given2d_suppl}.
The performance of the modified scheme is much worse than that of the original scheme proposed in the paper, which confirms the effectiveness of \ml{training a separate} 3D pose lifter for each joint group.

\begin{table*}
\caption{Performance comparison results on the Human3.6M data set in ``Case 1'' with ``Protocol 1''. `Single' means that we trained a single pose lifter to handle all of the joint groups.}
\centering
\resizebox{\textwidth}{!}{%
\begin{tabular}{l|rrrrrrrrrrrrrrr|r} \toprule
Protocol 1 & Direct. & Discuss & Eating & Greet & Phone & Photo & Pose & Purch. & Sitting & SittingD & Smoke & Wait & WalkD & Walk & WalkT & Avg \\ \toprule
Ours ($n_g=10$) & 33.9 & 40.5 & 34.8 & 38.6 & 40.4 & 50.2 & 41.0 & 39.8 & 55.4 & 48.8 & 39.4 & 40.4 & 42.3 & 36.5 & 31.9 & 40.9 \\
Ours ($n_g=10$, Single)  & 132.1 & 155.2 & 172.4 & 147.2 & 182.5 & 176.2 & 126.3 & 183.3 & 225.8 & 195.7 & 165.3 & 145.6 & 174.2 & 149.3 & 147.5 & 165.9 \\
Ours ($n_g=13$) & 34.5 & 41.7 & 35.4 & 39.4 & 41.2 & 50.9 & 41.4 & 40.9 & 56.0 & 47.8 & 40.0 & 41.1 & 43.7 & 35.6 & 32.6 & 41.5 \\
Ours ($n_g=13$, Single)  & 94.4 & 106.5 & 127.0 & 107.2 & 123.3 & 130.4 & 93.0 & 119.9 & 157.7 & 147.2 & 118.0 & 102.7 & 123.6 & 102.0 & 94.9 & 117.1 \\
Ours ($n_g=15$) & 35.3 & 42.6 & 37.1 & 39.9 & 43.3 & 52.1 & 42.8 & 40.5 & 55.9 & 50.5 & 41.4 & 41.9 & 44.3 & 36.0 & 32.4 & 42.6 \\
Ours ($n_g=15$, Single)  & 82.2 & 82.0 & 88.6 & 92.7 & 78.8 & 112.0 & 77.5 & 87.5 & 107.3 & 91.9 & 79.4 & 80.8 & 93.5 & 85.0 & 76.6 & 86.7 \\
\bottomrule
\end{tabular}%
}
\label{tab:given2d_suppl}
\end{table*}

\subsection{Bias analysis of \sh[1]{the} domain conversion layer}\label{sec:domainconversion}
We have proposed a novel domain conversion layer which transforms \ml{a heatmap} to its 2D coordinate\ght[1]{s} in the paper.
The proposed domain conversion layer consists of differentiable linear operations, which allows the back-propagation of gradients in the training process.
However, there might be a bias between the output of the domain conversion layer and that of the corresponding argmax layer.
In \shr[1]{the}{an} ideal case that the \ghr[1]{heatmap distribution \sht[1]{is }symmetric with respect to a joint position, which is the case of ground truth heatmaps}{mean and the peak of the heatmap distribution coincide}, we can verify that there is no bias between the two.\mlc{Oh, I discovered that a symmetric distribution is not enough for having zero bias. For example, think of a donut-like distribution. To be precise, there is no bias when the peak and the mean of a distribution is equal. So we have to modify the previous sentence.}
\mlr{Note here that the normalized ground truth heatmap distribution}{Note that \ght[1]{the heatmap of }a 2D pose estimator is trained based on a ground truth heatmap, of which its normalized distribution is a Gaussian distribution. Let us assume that, in an ideal case, the actual (normalized) output heatmap} of the $i$th joint is a Gaussian distribution which is described as
\begin{equation}
\begin{split}
	\big[\tilde{h}_i(I_k)\big]_{(p,q)} &= \frac{\textrm{exp}\left(-\frac{1}{2\gh[1]{\sigma^2}}\left((p-x_i)^2+(q-y_i)^2)\right)\right)}{\sqrt{(2\pi \gh[1]{\sigma^2})^2}} \triangleq \mathbf{N}_{x,y}(p,q)   \\
    &=\frac{\textrm{exp}\left(-\frac{1}{2\gh[1]{\sigma^2}}\left(p-x_i\right)^2\right)}{\sqrt{2\pi \gh[1]{\sigma^2}}}\cdot \frac{\textrm{exp}\left(-\frac{1}{2\gh[1]{\sigma^2}}\left(q-y_i\right)^2\right)}{\sqrt{2\pi \gh[1]{\sigma^2}}} \\
    &\triangleq \mathbf{N}_x(p)\mathbf{N}_y(q),
\end{split}
\end{equation}
where $x_i$ and $y_i$ are the 2D coordinate\ml{s} of the $i$th joint, and \ml{$\sigma^2$} is \ml{the} variance of the Gaussian distribution.
We can \shr[1]{verity}{verify} that the proposed domain conversion layer provides \mlt{the }\sh[1]{an} unbiased 2D coordinate\ml{s} from its ground truth heatmap as
\begin{equation}
\begin{split}
	&\sum_p \sum_q \big[\tilde{h}_i(I_k)\odot ([1:l_x] \otimes \mb{1}^T)\big]_{(p,q)} = \sum_p \sum_q \big[\tilde{h}_i(I_k)\big]_{(p,q)} \cdot q \\
    &= \sum_p \sum_q \mathbf{N}_{x,y}(p,q) \cdot q = \sum_p \sum_q \mathbf{N}_x(p)\mathbf{N}_y(q) \cdot q \\
    &= \sum_p \mathbf{N}_x(p) \sum_q \mathbf{N}_y(q)\cdot q = x_i.
\end{split}
\end{equation}
We can verify \ml{a similar relation} for $y_i$\mlt{ based on \ml{a} similar derivation}.

\begin{figure*}
    \centering
    \includegraphics[height=12.0cm]{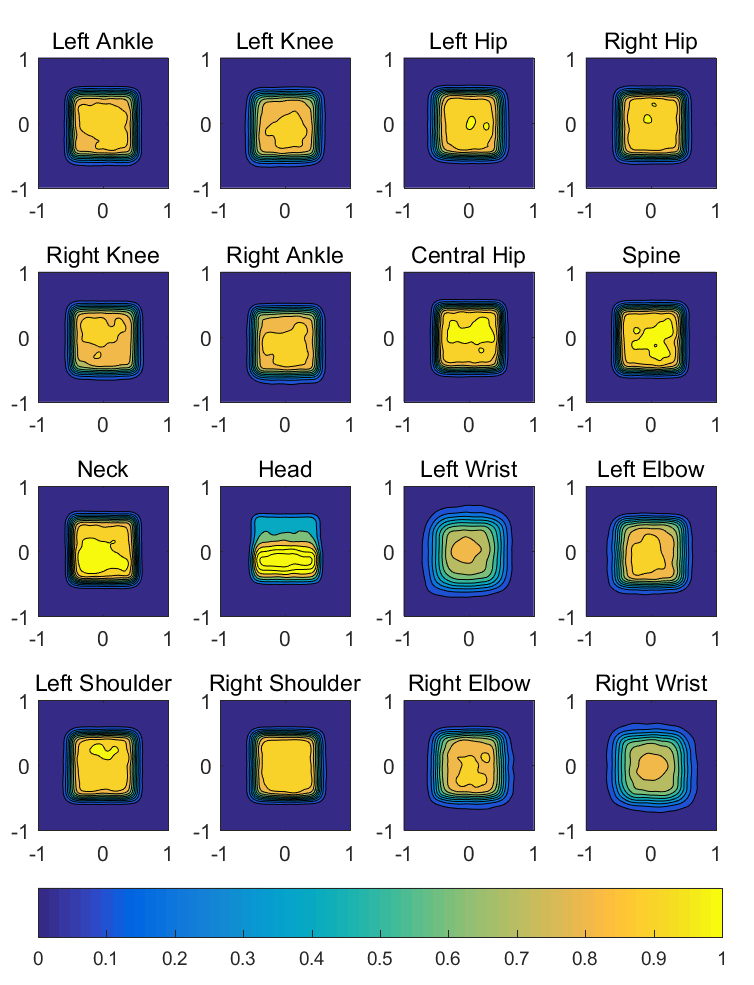}
    \caption{A visualization of the estimated probability density of the biases \ml{for} each joint.}
    \label{fig:bias}
\end{figure*}

\mlr{On the other hand}{In reality}, we can\mlt{ }not expect \mlr{a symmetric heatmap distribution}{to have such an ideal output distribution} from \ml{a} 2D pose estimator.
In this case, there might be a bias between the output of the proposed domain conversion layer and that of the corresponding argmax layer.
However, we empirically found out that there is only a slight difference.
We empirically calculated the biases on the test samples of the Human3.6M data set\ml{,} based on the proposed framework trained on the \mlr{Human3.6M}{same} data set.
In this process, we sub-sampled the test set by the factor of $5$ due to the large \mlr{number of samples}{sample size}, and the biases were evaluated on the $64\times 64$-size heatmaps.
\ml{Based on these bias samples, we performed nonparametric estimations to find their underlying} probability density \ml{for} each joint based on a normal kernel function.
The results are shown in Figure \ref{fig:bias}.
We can verify that the norms of the \mlt{most }biases are \ml{mostly} less than \ml{a} pixel, which is a slight difference considering the performance improvement based on the proposed end-to-end fine-tuning process.

\subsection{Qualitative examples}\label{sec:qual}

\begin{figure*}[t]
    \centering
    \includegraphics[height=17cm]{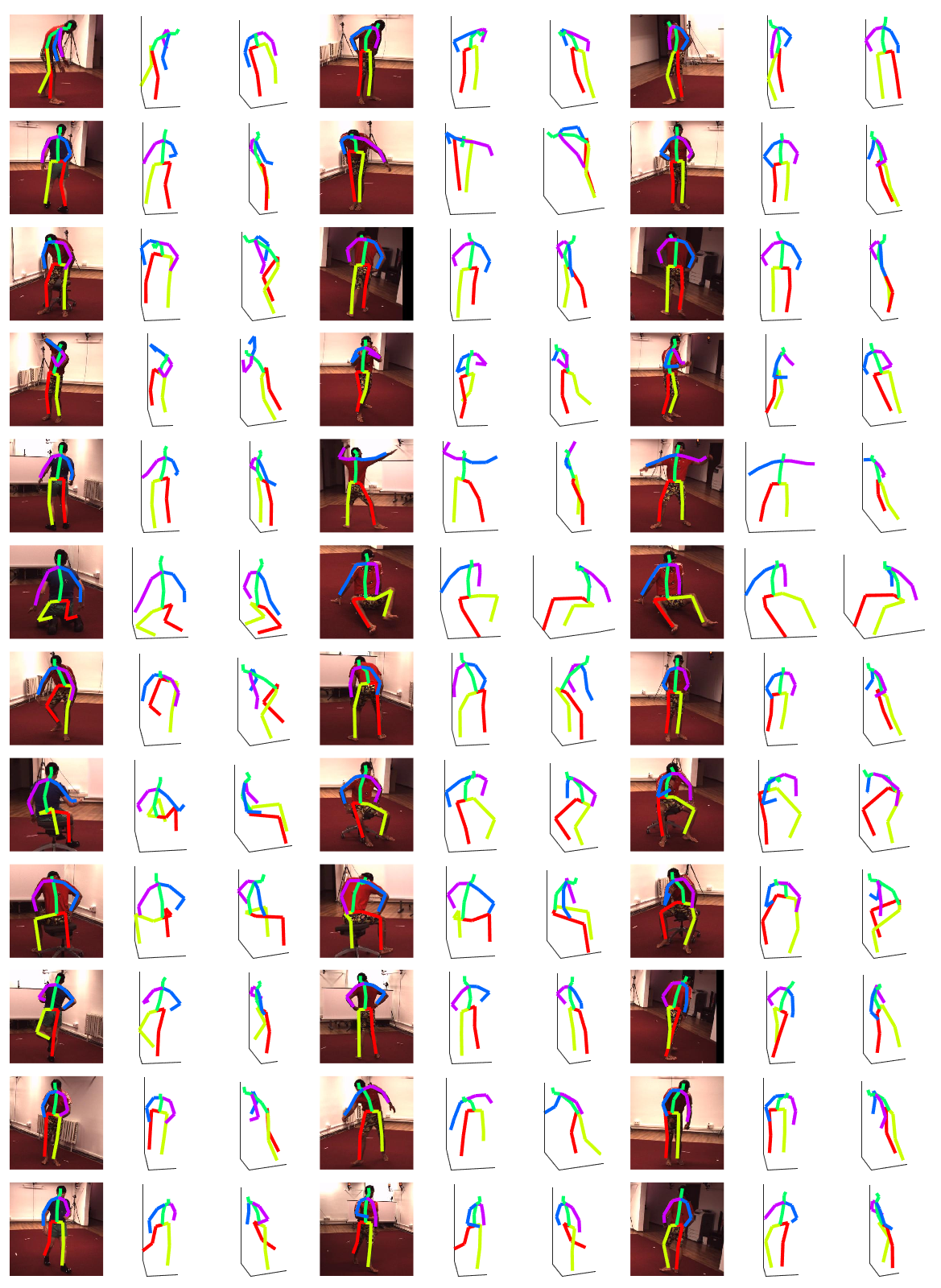}
    \caption{3D estimation \mlt{results of some }examples \mlr{in the test set}{for the test samples} of the Human3.6M data set. Left: Input RGB images and the corresponding 2D pose estimations. Middle, Right: 3D pose estimation results in two different views.}
    \label{fig:example_human36m_suppl}
\end{figure*}

\begin{figure*}[t]
    \centering
    \includegraphics[height=14.2cm]{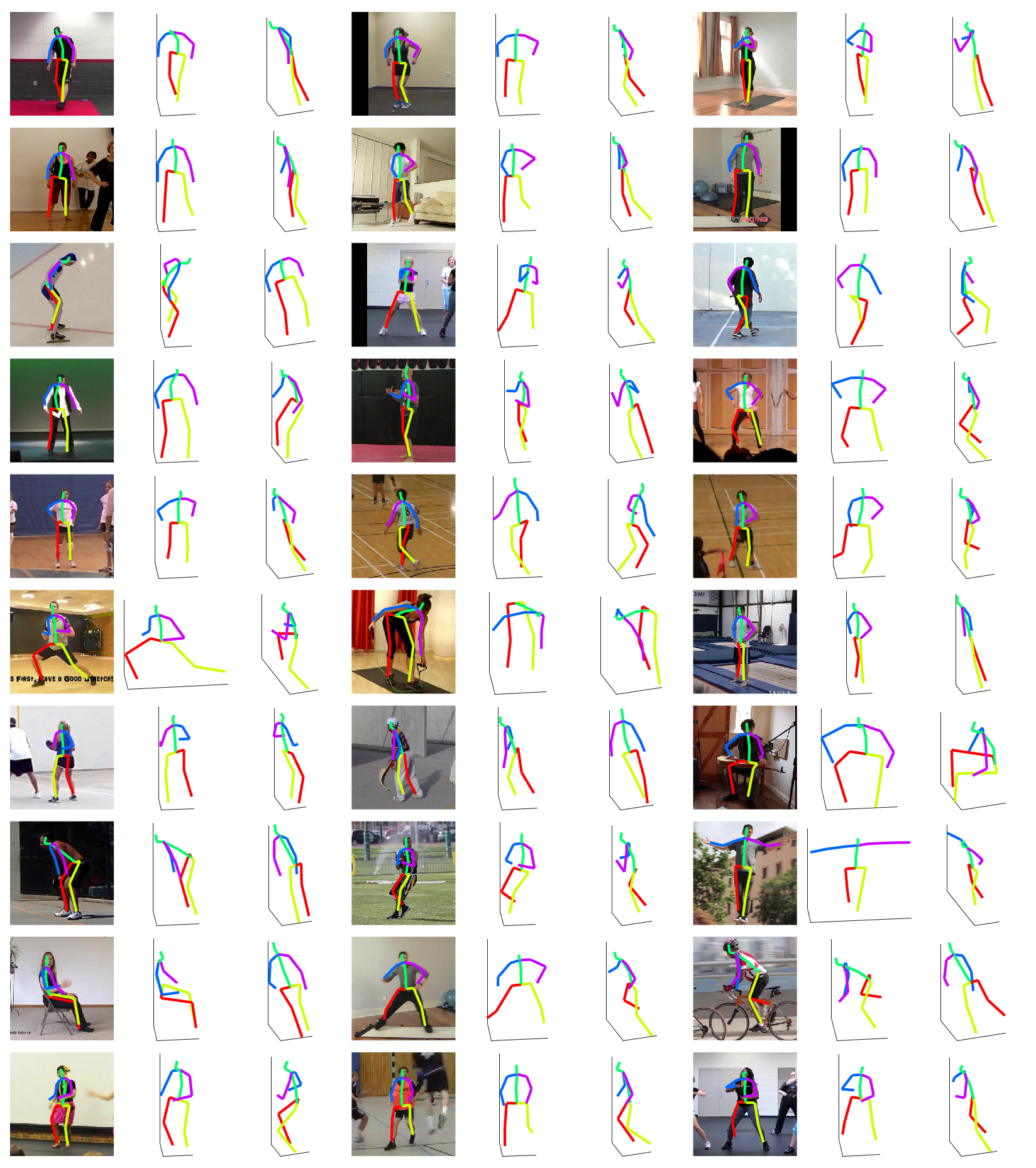}
    \caption{3D estimation \mlt{results of some }examples \mlr{in the test set}{for the test samples} of the MPII data set. Left: Input RGB images and the corresponding 2D pose estimations. Middle, Right: 3D pose estimation results in two different views.}
    \label{fig:example_mpii_suppl}
\end{figure*}

We present some more qualitative results in this section.
Some results on the Human3.6M data set are visualized in Figure \ref{fig:example_human36m_suppl}, and those on the MPII data set \cite{andriluka:mpii:cvpr14} are visualized in Figure \ref{fig:example_mpii_suppl}.
We can confirm that \ghr[1]{the estimation results are acceptable}{the proposed framework successfully estimates a 3D pose from an image}\mlc{This expression is too weak}.

\end{document}